\documentclass[floatfix,rmp,twocolumn,twoside]{revtex4}

\usepackage{amssymb}
\usepackage{amsmath}
\usepackage[english]{babel}
\usepackage{graphicx}

\makeatletter
\g@addto@macro\@verbatim\small
\makeatother

\bibpunct{[}{]}{,}{n}{}{,}

\begin{document}

\title{A Lexical Analysis Tool with Ambiguity Support}
\author{Luis~Quesada, Fernando~Berzal, and Francisco J.~Cortijo\\
  Department of Computer Science and Artificial Intelligence, CITIC, University of Granada, \\
  Granada 18071, Spain \\
  \textit{lquesada@decsai.ugr.es, fberzal@decsai.ugr.es, cb@decsai.ugr.es}
  }

\begin{abstract}
Lexical ambiguities naturally arise in languages. We present Lamb, a lexical analyzer that produces a lexical analysis graph describing all the possible sequences of tokens that can be found within the input string. Parsers can process such lexical analysis graphs and discard any sequence of tokens that does not produce a valid syntactic sentence, therefore performing, together with Lamb, a context-sensitive lexical analysis in lexically-ambiguous language specifications.
\end{abstract}

\maketitle

\section{Introduction}
\label{sec:introduction}
\noindent A lexical analyzer, also called lexer or scanner, is a piece of software that processes an input string conforming to a language specification and produces a sequence of the tokens or terminal symbols found in it. The obtained sequence of tokens is then usually fed to a parser or syntactic analyzer as the next step of a data translation, compilation or interpretation procedure.

Sometimes, lexical ambiguities may show up in a language specification. Lexical ambiguities occur when an input string simultaneously corresponds to several token sequences \cite{Nawrocki1991}.

The traditional way of choosing a sequence amongst potential alternatives \cite{Levine1992} involves assigning an unique priority to each token. This causes that, when the regular expressions associated to two different tokens match the same fragment of the input string, only the one with the greater priority will be considered.

However, the language developer may want similar substrings to be recognized as different sequences of tokens depending on their context. This cannot be achieved with the unique priority approximation.

Statistical lexical analyzers also exist \cite{Markov1971}. Although statistical models may perform well in context-sensitive scenarios, they require intensive training and, as token types are actually guessed, they do not formally guarantee that the obtained token sequence will be what the developer intended.

When it comes to programming languages, data specification languages, or limited natural languages scenarios, the syntactic rules are clear as to what should be accepted. The usage of statistical models introduces an unpredictable possibility of error during token recognition that would render scanning and parsing theoretically and pragmatically unfeasible.

Our proposal, Lamb (standing for \emph{Lexical AMBiguity}), performs a lexical analysis that efficiently captures all the possible sequences of tokens and generates a lexical analysis graph that describes them all. A subsequent parsing process discards any sequence of tokens that does not provide a valid syntactic sentence conforming to the syntactic rule set of the language specification. This solves the lexical ambiguity problem with formal correctness.

Therefore, Lamb allows language developers to specify more complex languages than traditional techniques. Token priorities are still supported but their usage is optional. Several tokens may be set to share the same priority if the developer wants ambiguities involving them to be considered.

As research in lexical analyzers sets the basis for the application of parsers, it inherits their application fields: the compilation or interpretation of source code written in programming languages \cite{Aho2006}, the interpretation and integration of data in data mining applications \cite{Han2006}, and natural language processing \cite{Jurafsky2008}.

\section{Background}
\noindent The IEEE POSIX P1003.2 standard describes the requirements of the \emph{lex} and \emph{yacc} tools \cite{Levine1992}, which are a traditional lexical analyzer generator and a traditional syntactic analyzer generator, respectively. Implementations of these tools are typically used in conjunction:

\begin{itemize}
\item \emph{Lex} generates a lexer that takes as input a set of token types, associated regular expressions \cite{Sipser2005}, and the string to be scanned; and produces the sequence of tokens found in the string.

\item \emph{Yacc} generates a parser that takes as input the sequence of tokens and a syntactic rule set; and produces a parse tree.
\end{itemize}

Regarding ambiguities, \emph{lex} enforces the assignment of unique priorities to each token. Indeed, tokens are tried and matched in the very same order they have been specified.

The order of efficiency of a \emph{lex}-generated lexical analyzer is $O(n)$, being $n$ the input string length.

The example \emph{lex} specification in Figure \ref{fig:lexspecification1} shows an example of implicitly reserved words, as the words ``true'', ``false'', ``if'', or ``while'' will not be considered identifiers, because they will match \emph{BOOLEAN}, \emph{IF}, or \emph{WHILE} tokens before reaching the regular expression for \emph{IDENTIFIER}. Therefore, it is not possible for \emph{lex} to consider these words as identifiers in some contexts, even if syntactic rules make clear whether occurrences of these words should be considered as identifiers or not.

Statistical models as Hidden Markov Models \cite{Markov1971,Rabiner1989,Ephraim2002}, Hierarchical Hidden Markov Models \cite{Fine1998}, or Maximum Entropy Markov Models \cite{McCallum2000} consider the existence of implicit relationships between words, symbols, or characters that are close together in strings.

These models need intensive corpus-based training and they produce results with associated implicit probabilities.

It should be noted that, even though they can perform well in natural language processing, their training requirement is impractical for programming or data representation languages, especially when the syntactic rules provide all the needed context information to unequivocally identify tokens. Furthermore, the results are prone to interpretation errors that would render the analysis unusable.

The semi-syntactic lexical analyzer proposed in \cite{Shyu1986} brings some of the context information found in the syntactic rule set to the deterministic finite automaton that perform the lexical analysis. Although this technique considers context information found in syntactic rules, it is not able to capture syntactic ambiguities for its further consideration, since the minimal automaton needed to do this is a non-deterministic finite automaton, which would have increased complexity of the algorithm. Therefore, if the lexical ambiguities may cause syntactic ambiguities or, in other words, there are several syntactic interpretations of the input string due to lexical ambiguities, a Shyu-like lexer would be unable to find them.

\section{Lamb}
\noindent In contrast to the aforementioned techniques, Lamb is able to recognize and capture lexical ambiguities.

Our proposed algorithm takes as input the string to be scanned and a list of tokens associated to their corresponding regular expressions. It produces, as output, a lexical analysis graph, in which each token is connected to its following and preceding tokens in the input sequence.

Our algorithm consists of two steps: the scanning step, which recognizes all the possible tokens in the input string; and the graph generation step, which computes the sets of preceding and following tokens for each token and builds the resulting lexical analysis graph.

\subsection{The Scanning Step}

\noindent The pseudocode for the scanning step is shown in Figure \ref{fig:codescan}.

\begin{figure}[tb]
\noindent
\begin{verbatim}
if           return(IF);
while        return(WHILE);
true|false   return(BOOLEAN);
[_a-zA-Z]+   return(IDENTIFIER);
\end{verbatim}
\caption{Example lex specification with implicitly reserved words (``true'', ``false'', ``if'', and ``while'').}
\label{fig:lexspecification1}
\end{figure}

\begin{figure}[tb]
\noindent
\begin{verbatim}
for i in 0..input.length()-1:
  prio = -1
  for each matcher m in matcherlist:
    if (prio >= m.prio || prio == -1) &&
        (prio != 0 && next[j] < i):
      match = m.match(input,i)
      if match != null:
        priority = matcher.priority
        if m.isignore==false:
          t = new token(
               id = id,
               type  = matcher.type,
               text  = match,
               start = i,
               end   = i+match.length()-1
             )
          tokenlist.add(t)
          id++
        min = i+match.length()-1
        for each matcher n in matcherlist:
          if n.next <= min && n.next >= i:
            min = n.next
          if n.next > m.next:
            n.next = i+match.length()-1
        if i >= min:
          min = i
        m.next = min
        for each matcher n in matcherlist:
          if n.prio > m.prio:
            n.next = min
\end{verbatim}
\caption{Pseudocode of the scanning step in our lexical analysis algorithm.}
\label{fig:codescan}
\end{figure}

Our algorithm receives an input string called \emph{input} and a list of matchers called \emph{matcherlist}. Each matcher consists of a regular expression and its corresponding \emph{match} method, a \emph{priority} value, and a \emph{next} value.

The \emph{match} method tries to perform a match given the input string and a starting position in it.

The \emph{priority} value specifies the matcher priority. The value 0 is reserved for ignored patterns, which are patterns that represent text that does not correspond to tokens. Then, priority values for relevant token start at 1, being the lower the value, the higher the priority. If two tokens share the same priority value, the lexer will capture both of them if they overlap due to lexical ambiguities. If two tokens have distinct priority values and they start at the same position in the input string, only the greater priority token will be considered.

The \emph{next} value specifies the position before the next string position a matcher will be tried at. It defaults to -1, so every matcher will be tried at the 0 position.

The \emph{prio} variable represents the last priority that has been matched in the current input position. Its value is -1 if no match has been made, 0 if an ignored element match has been made, and a higher value if any token of that specific priority has been identified.

The \emph{min} variable is computed in order to determine the next position the current matcher will be tried at, and its value is the minimum of either the ending position of the found token or the ending position of any tokens that end before it.

This algorithm step has a theoretical order of efficiency of $O(n^2 \cdot l)$, being $n$ the input string length and $l$ the number of matchers in the lexer.

\subsection{The Graph Generation Step}

The algorithm pictured in Figure \ref{fig:codegraph} goes through the identified token list in reverse order and efficiently computes the sets of preceding and following tokens for every token. 

\begin{figure}[tb]
\noindent
\begin{verbatim}
for i in tokenlist.size()-1..0:
  t = tokenlist[i]
  for j in i+1..tokenlist.size()-1:
    tc = tokenlist[j]
    if (tc.start > t.end &&
        (tc.prevstart==tc.start ||
         (tc.prevstart<tc.start &&
          tc.prevstart<t.end))):
      t.addfollowing(tc)
      tc.addpreceding(t)
      tc.prevstart = min(t.start,
                           tc.prevstart)
\end{verbatim}
\caption{Pseudocode of the graph generation step in our lexical analysis algorithm.}
\label{fig:codegraph}
\end{figure}

The sets of preceding and following tokens of the token $x$ are defined in Equation \ref{eq:nextprev}, being $a,b,c$ tokens and $x_{start}$ and $x_{end}$ the starting and ending positions of the token $x$ in the input string.

\begin{equation}
\begin{split}
b \in &FOLLOWING(a), a \in PRECEDING(b)\textrm{ iif } \\
& a_{end}<b_{start}\textrm{ \& } \nexists c, c_{start}>a_{end}, c_{end}<b_{start} 
\end{split}
\label{eq:nextprev}
\end{equation}

The \emph{prevstart} variable in the pseudocode avoids the need of iterating through the token list to find out if there is any token between two particular tokens, because it represent the starting position values of preceding tokens, given a certain token.

After the following and preceding sets have been computed for every token, any token whose preceding set is empty is added to the start token set of the lexical analysis graph.

The graph generation has a theoretical order of efficiency of $O(t^2)$, being $t$ the number of tokens found.  As $t \leq n \cdot l$, the theoretical order of efficiency of this step is $O(n^2 \cdot l^2)$.

Both scanning and graph generation steps together have an order of efficiency of $O(n^2 \cdot l^2)$.

\section{Comparison}
\noindent In order to perform a formal comparison of traditional techniques and Lamb, we have implemented a simple (and inefficient) proof of concept parser that supports ambiguities and allows a lexical analysis guided by a syntactic rule set. This parser returns as many parse trees as they can be obtained by applying a set of syntactic rules to a lexical analysis graph.

Its pseudocode is shown in Figure \ref{fig:codeparser}. It iteratively tries to match every rule starting from every existing token and following any possible token path, and it adds the newly found tokens to the list until no new tokens have been found in an iteration.

\begin{figure}[tb]
\noindent
\begin{verbatim}
do:
  flag = false
  for each rule r in rules:
    for each token t in tokenlist:
      matches = r.match(t)
      if matches.size() != 0:
        for each match m in matches:
          if !tokenlist.contains(m):
            tokenlist.add(m)
            if m is start symbol
              start.add(m)
            flag = true
while flag = true
\end{verbatim}
\caption{Pseudocode of the proof of concept parser supporting ambiguities.}
\label{fig:codeparser}
\end{figure}

\begin{figure*}[p]
\noindent
\hspace{-0.28cm}
\centering
\includegraphics[scale=0.445]{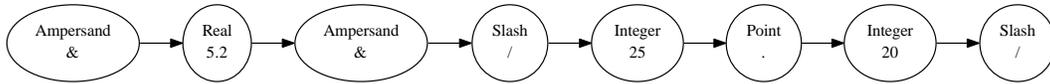}
\caption{Intended lexical analysis.}
\label{fig:e4}
\end{figure*}

\begin{figure*}[p]
\noindent
\hspace{-0.28cm}
\centering
\includegraphics[scale=0.445]{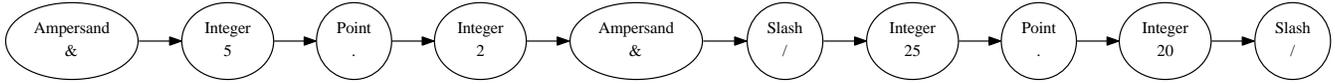}
\caption{Lexical analysis, as produced by \emph{lex}-alike lexers, when \emph{Integer} tokens are assigned greater priority than \emph{Real} tokens.}
\label{fig:e3}
\end{figure*}

\begin{figure*}[p]
\noindent
\hspace{-0.28cm}
\centering
\includegraphics[scale=0.445]{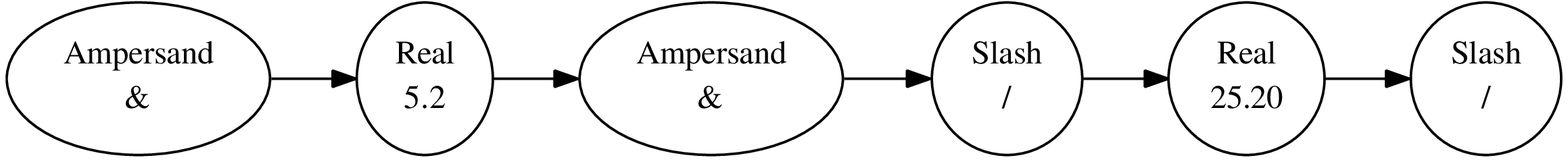}
\caption{Lexical analysis, as produced by \emph{lex}-alike lexers, when \emph{Real} tokens are assigned greater priority than \emph{Integer} tokens.}
\label{fig:e2}
\end{figure*}

\begin{figure*}[p]
\noindent
\hspace{-0.28cm}
\centering
\includegraphics[scale=0.445]{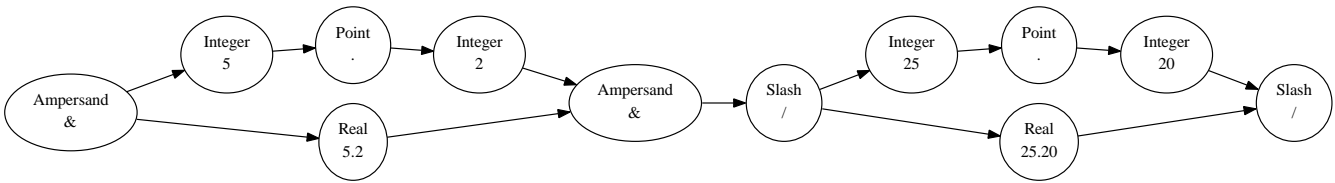}
\caption{Lexical analysis, as produced by Lamb, when \emph{Real} and \emph{Integer} tokens share priority value.}
\label{fig:e1}
\end{figure*}

\begin{figure*}[p]
\noindent
\hspace{-0.28cm}
\centering
\includegraphics[scale=0.445]{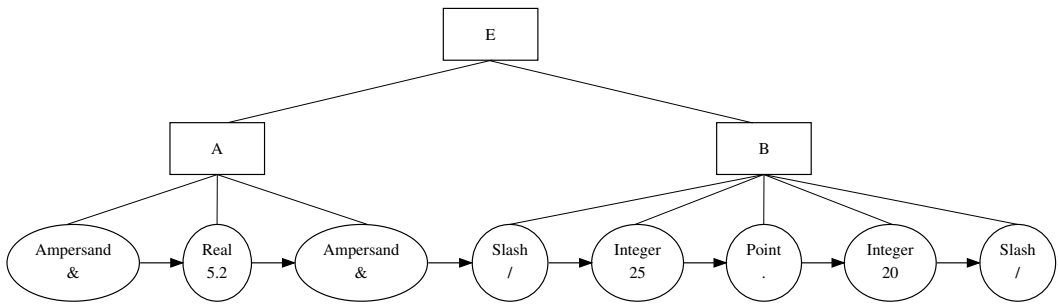}
\caption{Correct syntactic analysis produced by applying an ambiguity-supporting parsing technique to the lexical analysis graph produced by Lamb and shown in Figure \ref{fig:e1}.}
\label{fig:e5}
\end{figure*}

Given a language specification that describes the tokens listed in Figure \ref{fig:tokens}, the input string ``\&5.2\& /25.20/'' can correspond to the four different lexical analysis alternatives enumerated in Figure \ref{fig:analysis}, depending on whether the sequences of digits separated by points are considered real numbers or integer numbers separated by points.

\begin{figure}[htb!]
\noindent
\begin{verbatim}
   (-|\+)?[0-9]+              Integer
   (-|\+)?[0-9]+\.[0-9]+      Real
   \.                         Point
   \/                         Slash
   \&                         Ampersand
\end{verbatim}
\caption{Regular expressions and token names in the specification of our ambiguous language.}
\label{fig:tokens}
\end{figure}

\begin{figure}[htb!]
\noindent
\begin{itemize}
\item \texttt{Ampersand Integer Point Integer Ampersand Slash Integer Point Integer Slash}
\item \texttt{Ampersand Integer Point Integer Ampersand Slash Real Slash}
\item \texttt{Ampersand Real Ampersand Slash Integer Point Integer Slash}
\item \texttt{Ampersand Real Ampersand Slash Real Slash}
\end{itemize}
\caption{Different possible token sequences in an input string due to the lexically-ambiguous language specification in Figure \ref{fig:tokens}.}
\label{fig:analysis}
\end{figure}

\begin{figure}[htb!]
\noindent
\begin{verbatim}
E ::= A B
A ::= Ampersand Real Ampersand
B ::= Slash Integer Point Integer Slash
\end{verbatim}
\caption{Context-sensitive syntactic rules to resolve lexical ambiguities.}
\label{fig:srules}
\end{figure}

The syntactic rules shown in Figure \ref{fig:srules} illustrate a scenario of lexical ambiguity sensitivity. Depending on the surrounding tokens, which may be either \emph{Ampersand} tokens or \emph{Slash} tokens, the sequences of digits separated by points should be considered either \emph{Real} tokens or \emph{Integer Point Integer} token sequences. The expected results of analyzing the input string ``\&5.2\& /25.20/'' is shown in Figure \ref{fig:e4}.

In order to resolve the ambiguities when using a \emph{lex}-alike lexer, the developer can assign the \emph{Integer} token a greater priority than the \emph{Real} token. In that case, the only valid interpretation would be the one shown in Figure \ref{fig:e3}. The developer can also assign the \emph{Real} token a greater priority than the \emph{Integer} token. In that case, the only valid interpretation would be the one shown in Figure \ref{fig:e2}. Therefore, \emph{lex}-alike lexers cannot produce the token sequence that is needed to parse strings that belong to our language with lexical ambiguities.

Nonetheless, as Lamb is able to capture all the possible token sequences in the form of a lexical analysis graph, as shown in Figure \ref{fig:e1}, the later application of a parser supporting lexical ambiguities will produce the only possible valid sentence, which, in turn, is based on the only valid lexical analysis possible. Both of them are shown in Figure \ref{fig:e5}.

Even though statistical models as Hidden Markov Models may produce correct results in similar situations, they cannot be used for this kind of language specifications, where the specification states how each token is to be recognized. Besides, their results may not be always accurate, which difficults formally proving their correctness in a well-defined setting.

\section{Conclusions}
\noindent We have presented a lexical analyzer, Lamb, that supports lexical ambiguities. It performs a lexical analysis that efficiently captures all the possible sequences of tokens for lexically-ambiguous languages and it generates a lexical analysis graph that describes them all. Lamb supports assigning priorities to tokens as traditional techniques do but, in contrast to them, it does not enforce these priorities to be set and it allows for priority values to be shared. Tokens with shared priorities are considered valid alternatives instead of mutually-exclusive options.

The lexical graph can be then fed as input to a parser, which will discard any sequence of tokens that does not produce a valid syntactic analysis. In summary, our proposal performs a context-sensitive lexical analysis guided by syntactic rules and supports lexically-ambiguous language specifications.

\renewcommand{\baselinestretch}{0.98}
\bibliographystyle{plain}
{\small
\bibliography{doc}}
\renewcommand{\baselinestretch}{1}

\end{document}